\documentclass[10pt,twocolumn,letterpaper]{article}

\usepackage{cvpr}              

\usepackage{graphicx}
\usepackage{amsmath}
\usepackage{amssymb}
\usepackage{booktabs}
\usepackage{multirow}
\usepackage{algorithm}
\usepackage{algorithmic}
\usepackage{pifont}
\usepackage[accsupp]{axessibility}

\usepackage[pagebackref,breaklinks,colorlinks]{hyperref}

\usepackage[capitalize]{cleveref}
\crefname{section}{Sec.}{Secs.}
\Crefname{section}{Section}{Sections}
\Crefname{table}{Table}{Tables}
\crefname{table}{Tab.}{Tabs.}

\def\cvprPaperID{3975} 
\def\confName{CVPR}
\def\confYear{2023}

\begin{document}

\title{Reinforcement Learning-Based Black-Box Model Inversion Attacks}

\author{Gyojin Han ~~~~~~~~~~~~~ Jaehyun Choi ~~~~~~~~~~~~~ Haeil Lee ~~~~~~~~~~~~~ Junmo Kim\\
School of Electrical Engineering, KAIST\\
{\tt\small \{hangj0820, chlwogus, haeil.lee, junmo.kim\}@kaist.ac.kr}
}
\maketitle

\begin{abstract}
Model inversion attacks are a type of privacy attack that reconstructs private data used to train a machine learning model, solely by accessing the model. Recently, white-box model inversion attacks leveraging Generative Adversarial Networks (GANs) to distill knowledge from public datasets have been receiving great attention because of their excellent attack performance. On the other hand, current black-box model inversion attacks that utilize GANs suffer from issues such as being unable to guarantee the completion of the attack process within a predetermined number of query accesses or achieve the same level of performance as white-box attacks. To overcome these limitations, we propose a reinforcement learning-based black-box model inversion attack. We formulate the latent space search as a Markov Decision Process (MDP) problem and solve it with reinforcement learning. Our method utilizes the confidence scores of the generated images to provide rewards to an agent. Finally, the private data can be reconstructed using the latent vectors found by the agent trained in the MDP. The experiment results on various datasets and models demonstrate that our attack successfully recovers the private information of the target model by achieving state-of-the-art attack performance. We emphasize the importance of studies on privacy-preserving machine learning by proposing a more advanced black-box model inversion attack.
\end{abstract}

\section{Introduction}
\label{sec:intro}

With the rapid development of artificial intelligence, deep learning applications are emerging in various fields such as computer vision, healthcare, autonomous driving, and natural language processing.
As the number of cases requiring private data to train the deep learning models increases, the concern of private data leakage including sensitive personal information is rising.
In particular, studies on privacy attacks \cite{rigaki2020survey} show that personal information can be extracted from the trained models by malicious users.
One of the most representative privacy attacks on machine learning models is a model inversion attack, which reconstructs the training data of a target model with only access to the model. 
The model inversion attacks are divided into three categories, 1) white-box attacks, 2) black-box attacks, and 3) label-only attacks, depending on the amount of information of the target model. The white-box attacks can access all parameters of the model. The black-box attacks can access soft inference results consisting of confidence scores, and the label-only attacks only can access inference results in hard label forms.

The white-box model inversion attacks \cite{Zhang_2020_CVPR, Chen_2021_ICCV, NEURIPS2021_50a074e6} have succeeded in restoring high-quality private data including personal information by using Generative Adversarial Networks (GANs) \cite{NIPS2014_5ca3e9b1}. 
First, they train the GANs on separate public data to learn the general prior of private data.
Then benefiting from the accessibility of the parameters of the trained white-box models, they search and find latent vectors that represent data of specific labels with gradient-based optimization methods. 
However, these methods cannot be applied to machine learning services such as Amazon Rekognition \cite{Amazon} where the parameters of the model are protected.
To reconstruct private data from such services, studies on black-box and label-only model inversion attacks are required. Unlike the white-box attacks, these attacks require methods that can explore the latent space of the GANs in order to utilize them, as gradient-based optimizations are not possible.
The recently proposed Model Inversion for Deep Learning Network (MIRROR) \cite{an2022mirror} uses a genetic algorithm to search the latent space with confidence scores obtained from a black-box target model. 
In addition, Boundary-Repelling Model Inversion attack (BREP-MI) \cite{Kahla_2022_CVPR} has achieved success in the label-only setting by using a decision-based zeroth-order optimization algorithm for latent space search. 

Despite these attempts, each method has a significant issue. BREP-MI starts the process of latent space search from the first latent vector that generates an image classified as the target class. This does not guarantee how many query accesses will be required until the first latent vector is found by random sampling, and in the worst case, it may not be possible to start the search process for some target classes. In the case of MIRROR, it performs worse than the label-only attack BREP-MI, despite its use of confidence scores for the attack.
Therefore, we propose a new approach, Reinforcement Learning-based Black-box Model Inversion attack (RLB-MI), as a solution that is free from the aforementioned problems. We integrate reinforcement learning to obtain useful information for latent space exploration from the confidence scores.
More specifically, we formulate the exploration of the latent space in the GAN as a problem in Markov Decision Processes (MDP).
Then, we provide the agent with rewards based on the confidence scores of the generated images, and use update steps in the replay memory to enable the agent to approximate the environment including latent space.
Actions selected by the agent based on this information can navigate latent vectors more effectively than existing methods. 
Finally, we can reconstruct private data through the GAN from the latent vectors.
We experiment with our attack on various datasets and models. The attack performance is compared with various model inversion attacks in three categories.
The results demonstrate that the proposed attack can successfully recover meaningful information about private data by outperforming all other attacks.

\section{Related Works}
\label{sec:relatedworks}
\subsection{Model Inversion Attacks}

Model inversion attack is a type of privacy attack on machine learning models, which reconstructs data used for training.
Early works focused on the white-box condition where the trained model is fully accessible.
Fredrikson \etal demonstrated the severeness of model inversion attacks by extracting sensitive attributes \cite{10.5555/2671225.2671227} and facial images from low-complexity models \cite{10.1145/2810103.2813677}.
However, early white-box model inversion attacks had clear limitations by being unable to reconstruct high-dimensional data from complex models.

As a solution for the limitations, many white-box model inversion attacks utilized a generative model.
Zhang \etal \cite{Zhang_2020_CVPR} proposed a model inversion attack named Generative Model Inversion attack (GMI) using a GAN \cite{NIPS2014_5ca3e9b1} trained on public data.
They reconstructed images by searching the latent space of GAN rather than the image space. The manifold approximation ability of GAN effectively narrowed the search space of the attack model.
Knowledge-Enriched Distributional Model Inversion attack (KED-MI) \cite{Chen_2021_ICCV} improved upon GMI by using an inversion-specific GAN where the discriminator performs multi-class inferences.
The discriminator of the inversion-specific GAN can distinguish between real and fake data, as well as predict the outputs of the target network. It makes the inversion-specific GAN extract useful knowledge of the target model from public data.
In addition, KED-MI can create diverse images by extracting the distribution of the target class data, not just one training data of the target class.
Variational Model Inversion attack (VMI) \cite{NEURIPS2021_50a074e6} formulated the model inversion attacks as a variational inference problem with the framework containing the deep normalizing flows and styleGAN \cite{Karras_2019_CVPR} framework. 

Unfortunately, the white-box attacks are not suitable for situations where there is no information about the parameters of the trained model. Thereby, black-box model inversion attacks requiring only access to soft labels were proposed. 
Learning-Based Model Inversion attack (LB-MI) \cite{yang2019adversarial} used a structure similar to an auto-encoder to train an inversion model that behaves in a reverse way to the target network. The inversion model requires access to not only the queries constructed by the attacker but also confidence scores for the user-entered data that is the target of inversion.
Salem \etal \cite{247690} presented an attack that leaks information of trained data used for an update by computing the difference between outputs before and after the model update. Both methods have a limitation in that they require information that is generally difficult for attackers to access. Model Inversion for Deep Learning Network (MIRROR) \cite{an2022mirror} showed that GANs can be utilized in black-box model inversion attacks using genetic algorithms.
Additionally, a label-only model inversion attack, Boundary-Repelling Model Inversion attack (BREP-MI) \cite{Kahla_2022_CVPR}, has been proposed. BREP-MI uses only hard labels in the inversion process.
BREP-MI estimates a gradient from the labels of the points on the sphere centered on the current latent vector. The latent vector is updated using the estimated gradient so that a GAN reconstructs a representative image of the target class from the latent vector. 

\begin{figure*}[ht]
\begin{center}
\includegraphics[width=17.25cm]{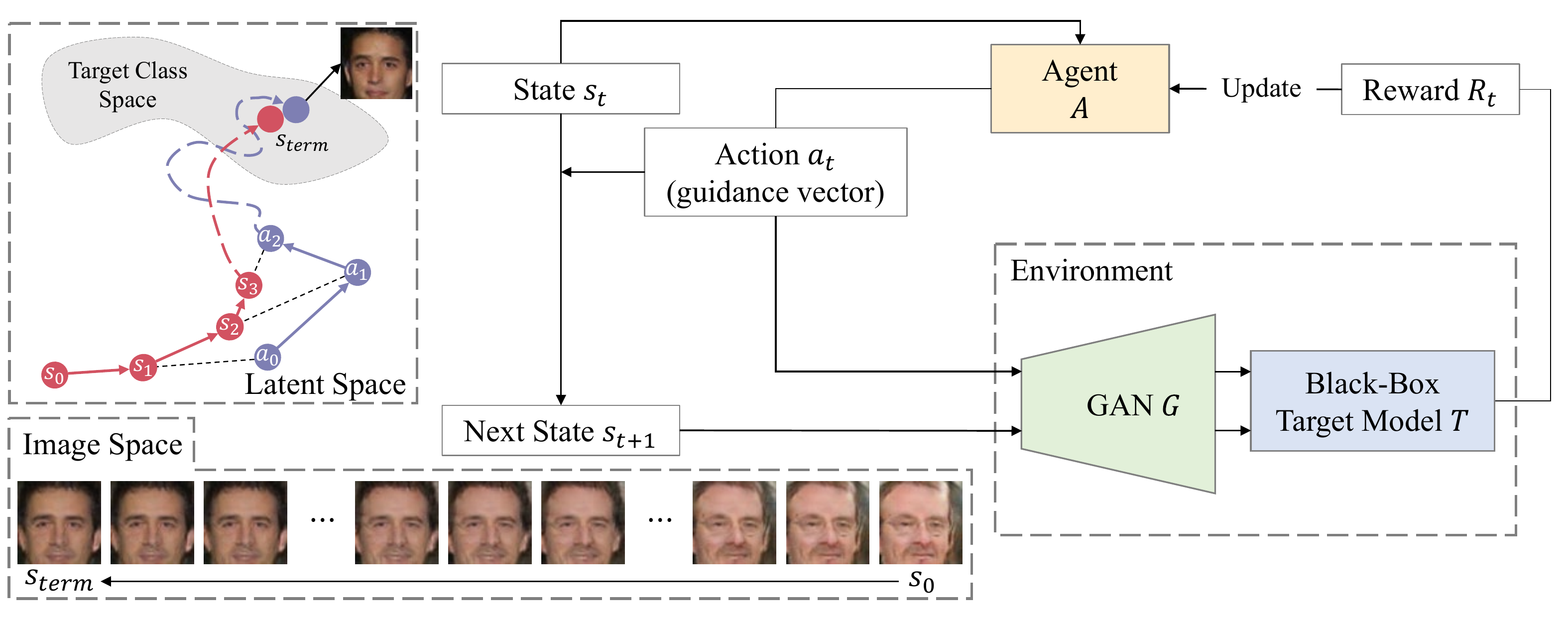}
\caption{Overview of the latent space search formulated as an MDP problem and the training process of a reinforcement learning agent. In the latent space, the state $s_t$ moves in the direction of the action $a_t$ by the distance determined with the diversity factor $\alpha$ for each step. $G$ generates images from the updated state $s_{t+1}$ and action $a_t$ to provide a reward to the agent $A$ via target model $T$. In addition, we visualized the state change from the initial state $s_0$ to the terminal state $s_{term}$ within one episode at the image space.}
\label{main_figure}
\end{center}
\end{figure*}

\subsection{Deep Reinforcement Learning}

Studies on reinforcement learning (RL) using deep neural networks began to draw attention as Deep Q-Network (DQN) \cite{mnih2015human} showed comparable performance to human experts in Atari 2600 games \cite{bellemare2013arcade}.
DQN successfully solves tasks with high-dimensional state inputs, such as raw pixels, by using deep neural networks to estimate the action-value function.
However, DQN cannot be immediately applied to problems with high-dimensional, continuous action spaces because it works by finding an action that maximizes the value of the action-value function.
To solve the problems with high-dimensional, continuous action spaces, Deep Deterministic Policy Gradient (DDPG) \cite{DBLP:journals/corr/LillicrapHPHETS15} was proposed.
DDPG is a model-free and off-policy algorithm using an actor-critic approach based on Deep Policy Gradient (DPG) \cite{pmlr-v32-silver14}.
It stabilized learning by applying DQN's idea of replay buffer and target networks to an actor-critic approach. Even after DDPG, many deep reinforcement learning methods have been proposed to improve DDPG. To overcome the overestimation bias problem, Twin Delayed Deep Deterministic Policy Gradient (TD3) \cite{pmlr-v80-fujimoto18a} algorithm is proposed. TD3 solves the problem by using some tricks like clipped double Q-learning and delayed policy updates. In addition, Soft Actor-Critic (SAC) \cite{pmlr-v80-haarnoja18b} was proposed to deal with problems with complex environments. SAC showed substantial improvement in exploration and stability by adding an entropy maximization term to the standard maximum reward reinforcement learning objective. 

\section{Proposed Method}

In this section, we will present our method, Reinforcement Learning-based Black-box Model Inversion attacks (RLB-MI). The overview of RLB-MI is illustrated in \Cref{main_figure}.
\subsection{Problem Formulation}
\textbf{Attacker's goal.} 
The goal of the model inversion attacks is to reconstruct representative data of a target class $y$ from a target model $T$ trained with a private dataset $D_{pvt}$.
The target model $T:\mathbf{x}\rightarrow{[0,1]^K}$ has learned the mapping from an image $\mathbf{x}\in\mathbb{R}^d$ to a label where K is the number of $D_{pvt}$'s classes and $d$ is the dimension of the input image. 

\textbf{Attacker's knowledge.}
As our method deals with the black-box setting, the attacker can only access queries organized with the data entered by the attacker and the soft label corresponding to the data.
Additionally, the attacker is aware of the purpose of the target model. As mentioned in the previous studies \cite{Zhang_2020_CVPR, Chen_2021_ICCV, NEURIPS2021_50a074e6, an2022mirror, Kahla_2022_CVPR}, the information on the task of the provided model or service is not only available but also can be easily inferred from the classes of the output. Based on this knowledge of the task, the attacker can access public datasets of the corresponding task.

\textbf{Overview.} Given the black-box model $T$ trained on the private dataset $D_{pvt}$, the black-box model inversion attacks aim to reconstruct $D_{pvt}$.
In parallel with recent model inversion attacks~\cite{Zhang_2020_CVPR, Chen_2021_ICCV, NEURIPS2021_50a074e6, an2022mirror, Kahla_2022_CVPR}, a GAN $G$ trained with the public dataset $D_{pub}$ is employed in our method.
The attacker in the black-box condition has no access to the structure and parameters of $T$ but the soft labels consisting of confidence scores.
Therefore, the main task of our method is searching latent vectors that generate images with high confidence scores for $D_{pvt}$'s classes from the latent space of $G$.
For the task, we approach and formulate the latent space search problem as a Markov Decision Process (MDP), and we apply reinforcement learning as it is the well-known solution for MDP with 1) an unknown environment, 2) continuous space, and 3) high-dimensional space. 
More specifically, we define the state space of MDP as the latent space of $G$. Then the state $s_t$ of each step $t$ has the same form as a latent vector.
The state $s_t$ is guided by an action $a_t$ named a guidance vector and updated to the next state $s_{t+1}$ as shown in \Cref{main_figure}.
Finally, the reward is formulated with the confidence scores of the images generated from $s_{t+1}$ and $a_t$.

\subsection{MDP for Latent Space Search} 

We describe the components of the MDP for latent space search: state, action, state transition, and reward.

\textbf{State.} The state space of this MDP is the latent space of $G$. For each episode, the first state $s_0$ is a $k$-dimensional standard normal random vector:
\begin{equation}
s_{0}\sim\mathcal{N}_k(\mathbf{0},\mathbf{1}), \quad s_0\in\mathbb{R}^{k},
\end{equation}
where $k$ is the dimension of the latent space.
At every step $t$, the state $s_t$ is updated by the action $a_t$.

\textbf{Action.}
We want the actions to guide the random initial latent vector to the highly rewarding final latent vector. In a broad sense, we perceive this problem as a reinforcement learning-based path-finding problem. In conventional path-finding problems, actions are defined as displacements $\Delta{s}$ from the current state $s_t$ to the next state $s_{t+1}$.
However, unlike the bounded two-dimensional space in path-finding, the latent space is an unrestricted high-dimensional space. When the action is defined as a displacement vector in our formulation, a reinforcement learning agent cannot converge and reach local minima or total failure due to a large state variance and a narrow exploration area compared to the latent space.
Therefore, we consider the action space as the whole latent space. We define a latent vector-shaped action as a guidance vector.
As can be seen in \Cref{main_figure}, the action is selected in the same space as the state by the definition. This enables a wide exploration of the entire latent space, preventing the agent from falling to local minima and guaranteeing convergence of the agent.
The action $a_t$ for the state $s_t$ in each step $t$ is determined by a reinforcement agent $A:\mathbb{R}^{k}\rightarrow{\mathbb{R}^{k}}$ :
\begin{equation}
a_t = A(s_t), \quad a_t\in\mathbb{R}^{k}.
\end{equation}
The state transition through the action is defined below.

\textbf{State transition.} 
We update the state by moving the state toward the action at each step with a diversity factor $\alpha$ which is used as a weight for the current state during state transition to determine the movement distance.
The state transition at step $t$ is as follows:
\begin{equation}
s_{t+1} = \alpha \cdot s_t + (1 - \alpha) \cdot a_t.
\end{equation}
The reason we named $\alpha$ as a diversity factor is that we present $\alpha$ as a hyperparameter that allows us to control the diversity of generated images. 
As mentioned by Wang \etal \cite{NEURIPS2021_50a074e6}, model inversion attacks have a trade-off between accuracy and diversity when reconstructing images. We can adjust the trade-off between accuracy and diversity using $\alpha$. The higher the $\alpha$, the greater the influence of the random initial latent vector on each episode and the more focused the agent is on generating images with high diversity. For example, if the value of $\alpha$ is zero, the next state will be the same as the action in the current step, therefore the agent simply focuses on generating one image with the highest probability for the target classifier. On the other hand, if the influence of the random initial latent vector increases as $\alpha$ grows, the agent is trained to find a variety of images with high probabilities for the target class.

\textbf{Reward.} After an update of the state through the action, the agent receives a reward from the environment. $G$ generates an image from the updated latent vector, and we can obtain a confidence score for the target class $y$ of the image by inference using the target network $T$. 
Since the action guides the movement direction of the state, the image generated from the action should also be close to the target class space. To place the state and action near the target class space, we need to provide a higher reward as the state and action have higher confidence scores.
Therefore, we compose a reward with a state score and an action score which are calculated as the logarithmic values of the confidence scores of the images created by each vector. The scores are calculated as follows:
\begin{equation}
\begin{split}
& r_1 = \log[T_y(G(s_{t+1}))] \\
& r_2 = \log[T_y(G(a_t))],
\end{split}
\end{equation}
where $r_1$ is the state score, and $r_2$ is the action score.
In addition, we want the reconstructed image to have features of the target class that distinguish it from images of other classes. Therefore, we propose an additional term $r_3$ that penalizes high confidence scores for other classes of images. 
We calculate the difference between the confidence score of the target class and the maximum confidence score of the other classes. As with previous scores, a logarithm is applied to the calculated value.
Since the logarithmic value for a number less than or equal to zero is undefined, the logarithm is taken to the greater of the subtracted value and a small positive number $\epsilon$.
The term $r_3$ is expressed as the formula:
\begin{equation}
r_3 = \log[\max\{\epsilon, T_y(G(s_{t+1})) - \max_{i \neq y}T_{i}(G(s_{t+1}))\}].
\end{equation}
The overall reward at each step $R_t$ is calculated as follows:
\begin{equation}
R_t = w_1 \cdot r_1 + w_2 \cdot r_2 + w_3 \cdot  r_3,
\end{equation}
where $w_n$ is the weight for the score $r_n$.

\subsection{Solving the MDP With Reinforcement Learning}

We solve the proposed latent space search problem formulated as the MDP through reinforcement learning.
Since the environment of the proposed MDP consisting of $G$ and $T$ is very complex, we require a robust reinforcement learning agent in complex environments.
In addition, we need an agent which can handle the continuous action space, because the action space in the MDP is defined as the latent space of $G$. 
Therefore, we solve the MDP using Soft Actor-Critic (SAC) \cite{pmlr-v80-haarnoja18b} that satisfies all of the mentioned points.
We train a SAC agent to select the appropriate action from a given state.
After the training, we can obtain a reconstructed image for each episode by providing a random initial vector to the trained agent as the initial state.
The entire procedure of agent training is illustrated in \Cref{alg:agent}.

\begin{algorithm}[!tb]
\caption{Agent training}
\label{alg:agent}
\textbf{Input}: GAN trained with public data $G$, Target classifier $T$, Target class $y$\\
\textbf{Output}: Learned agent $A$
\begin{algorithmic}[1] 
\STATE Initialize new agent $A$.
\FOR {$episode=1$ to $max\_episodes$}
    \STATE Initialize $s$ as a random latent vector.
    \FOR {$step=1$ to $max\_step$}
        \STATE $a \gets A(s)$
        \STATE $next\_s \gets \alpha \cdot s + (1 - \alpha) \cdot a$
        \STATE $r_1 \gets \log[T_y(G(next\_s))]$
        \STATE $r_2 \gets \log[T_y(G(a))]$
        \STATE $r_3 \gets \log[\max\{\epsilon, T_y(G(next\_s))$ \\
        \hskip 80pt $- \max_{i \neq y}T_{i}(G(next\_s))\}]$
        \STATE $reward \gets w_1 \cdot r_1 + w_2 \cdot r_2 + w_3 \cdot r_3$
        \IF {$step$ is $max\_step$}
            \STATE $done \gets True$
        \ELSE
            \STATE $done \gets False$
        \ENDIF
        \STATE $A.observe(s, a, reward, next\_s, done)$
        \STATE $A.update\_policy()$
        \STATE $s \gets  next\_s$
    \ENDFOR
\ENDFOR
\STATE \textbf{return} $A$
\end{algorithmic}
\end{algorithm}

\section{Experiments}

\subsection{Experimental Setting}

\textbf{Datasets.}
We evaluate our attack against target classifiers trained with representative face datasets, CelebFaces Attributes Dataset (CelebA) \cite{Liu_2015_ICCV}, FaceScrub Dataset \cite{Ng2014ADA}, and PubFig83 Dataset \cite{5981788}.
We split each dataset into a private dataset for training target classifiers and a public dataset for training a generative model.
There is no class intersection between the public dataset and the private dataset, therefore the generative model cannot learn class-specific information of the target classifier.

For CelebA, the private dataset consists of 30,027 images of 1,000 identities, and the public dataset consists of randomly selected 30,000 images from the remaining 9,177 identities as in previous studies \cite{Chen_2021_ICCV, Kahla_2022_CVPR}. For FaceScrub, among the total of 530 identities, all images of randomly selected 200 identities are used as a private dataset, and all images of the remaining 330 identities are used as a public dataset. For PubFig83, among the total of 83 identities, all images of randomly selected 50 identities are used as a private dataset, and all images of the remaining 33 identities are used as a public dataset.
In addition, we conduct experiments using Flickr-Faces-HQ Dataset (FFHQ) \cite{Karras_2019_CVPR} as a public dataset considering the situations where there are distributional shifts between the public dataset and private dataset.
In these experiments, 30,000 randomly selected images from FFHQ are used as public datasets.
All face images are center cropped and then resized to 64 $\times$ 64.

\textbf{Models.}
We experiment with the attacks on several popular network structures for fair comparisons.
Similar to previous studies \cite{Zhang_2020_CVPR, Chen_2021_ICCV, Kahla_2022_CVPR}, we use three network structures, VGG16 \cite{DBLP:journals/corr/SimonyanZ14a}, ResNet-152 \cite{He_2016_CVPR}, and Face.evoLVe \cite{Cheng_2017_ICCV}, for experiments.

\textbf{Implementation details.}
Hyperparameters for training target classifiers and GANs were set following the previous studies \cite{Zhang_2020_CVPR, Chen_2021_ICCV, Kahla_2022_CVPR}.
The target classifiers were trained using SGD (learning rate $0.01$, batch size $64$, momentum $0.9$, weight decay $1\times10^{-4}$) for 50 epochs and the best models were selected based on test accuracy. 
The GANs were trained using Adam (learning rate 0.004, batch size $64$, $\beta_1 = 0.5$, $\beta_2 = 0.999$) for 300 epochs. The SAC agents were trained using Adam with discount factor $\gamma = 0.99$, soft-update factor $\tau = 0.01$, learning rate $5\times10^{-4}$, replay memory size $1\times10^{6}$, batch size $256$, maximum step per one episode $1$, and diversity 
factor $\alpha = 0$ for 40,000 episodes. 
We set the reward weights $w_1$, $w_2$, and $w_3$ to $2$, $2$, and $8$, respectively.
The value of $\epsilon$ in these experiments is $1\times10^{-7}$.

\textbf{Evaluation metrics.} Zhang \etal \cite{Zhang_2020_CVPR} proposed metrics that can quantitatively evaluate model inversion attacks, unlike the previous qualitative evaluations that rely on visual inspection.
We briefly describe the evaluation metrics, \emph{attack accuracy}, \emph{K-nearest neighbor distance (KNN Dist)}, and \emph{feature distance (Feat Dist)}.

\emph{Attack Accuracy}: To evaluate the attack accuracy of reconstructed images, we train evaluation classifiers on private datasets. The evaluation classifiers must be different from the target classifiers because the reconstructed images can overfit the target classifiers. We use an architecture proposed by Cheng \etal~\cite{Cheng_2017_ICCV} for the evaluation classifiers. We fine-tuned a classifier pre-trained with MS-Celeb-1M \cite{guo2016ms} on private datasets, achieving test accuracy of 98\%, 99\%, and 96\% for PubFig83, FaceScrub, and CelebA datasets, respectively.

\emph{KNN Dist}: The KNN Dist is a metric to measure the average $L_2$ distance between the feature of the reconstructed image and the feature of the closest sample among the target label images. The feature was taken just before the fully connected layer of the evaluation classifier.

\emph{Feat Dist}: The Feat Dist is a metric to measure the $L_2$ distance between the feature of the reconstructed image and the centroid of features of the target label images. The feature was taken just before the fully connected layer of the evaluation classifier.

\subsection{Experimental Results}

\textbf{Baselines.} We compare the performance of our attack with representative model inversion attacks included in the three categories: white-box attacks, black-box attacks, and label-only attacks. The white-box attack baselines are GMI \cite{Zhang_2020_CVPR} and KED-MI \cite{Chen_2021_ICCV}. 
GMI is the first model inversion attack using GANs, and KED-MI shows the highest attack performance among the white-box attacks. 
The black-box and label-only baselines are LB-MI \cite{yang2019adversarial}, MIRROR \cite{an2022mirror}, and BREP-MI \cite{Kahla_2022_CVPR}. 
For fair comparisons, the same GAN was used for the attacks except for LB-MI and KED-MI as LB-MI does not use a GAN, and KED-MI uses its own generative model named the inversion-specific GAN.
We generated images using agents for each episode and selected the images with the highest confidence score for the target classifiers.
\begin{figure}[t]
\begin{center}
\centerline{
\includegraphics[width=1.0\columnwidth]{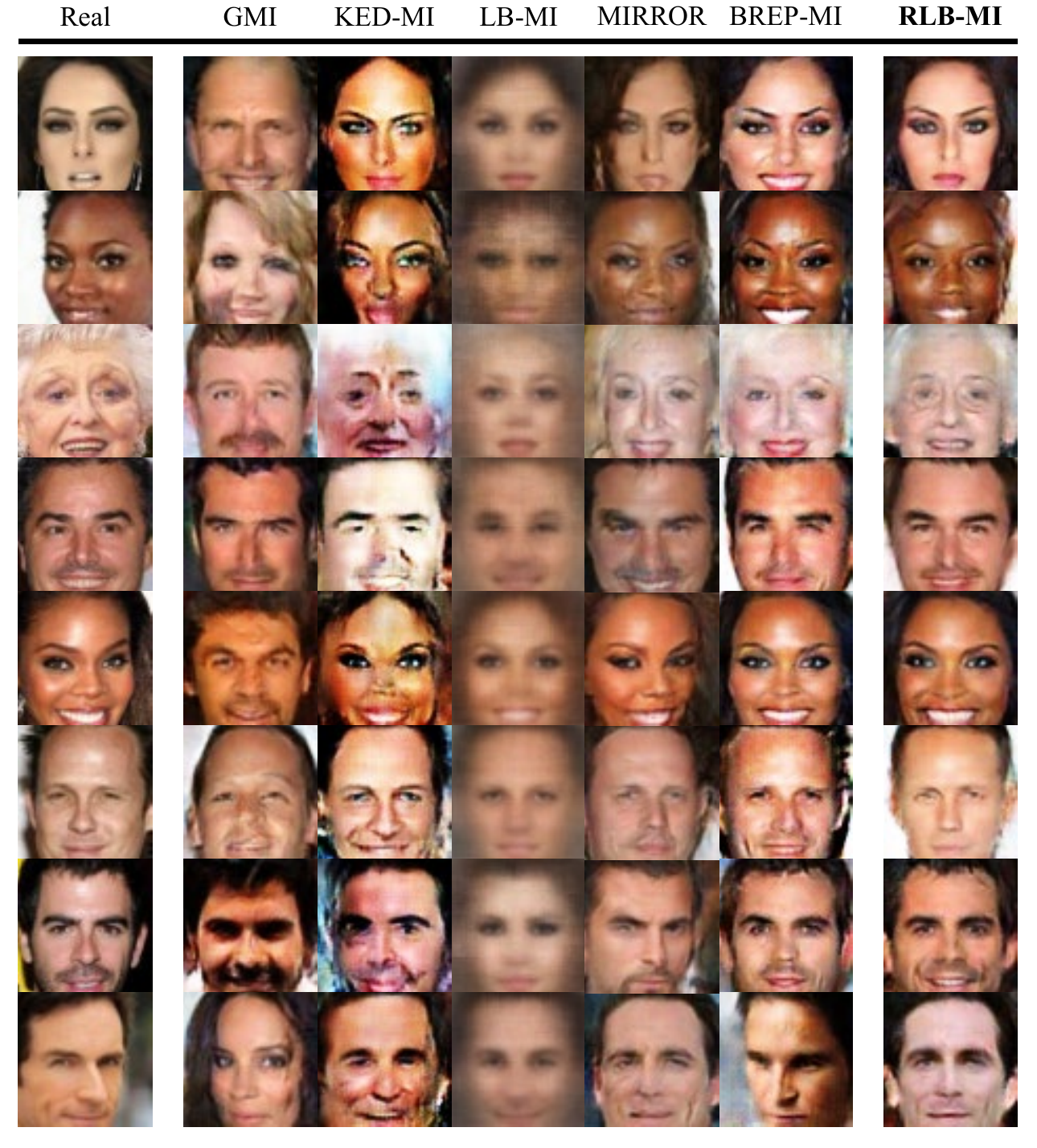}
}
\caption{Real samples and images reconstructed with model inversion attacks. The leftmost images are real samples and the images in the same row are for the same identity.}
\label{result-image}
\end{center}
\end{figure}
\begin{table}[t]
\centering
\begin{scriptsize}
\resizebox{\columnwidth}{!}{
\begin{tabular}{c|c|c|rrr}
  \toprule
 Model & Type & Method & Attack Acc & KNN Dist & Feat Dist \\  \midrule[0.7pt]
\multirow{6}{*}{\shortstack{VGG16 \\ (88\%)}} & \multirow{2}{*}{White-box} & GMI & 0.185 &  1693.7 & 1615.7 \\ 
   &  & KED-MI & \textbf{0.703} & 1334.0 & 1243.7 \\ 
   \cmidrule{2-6}
   &  \multirow{3}{*}{Black-box}& LB-MI & 0.075 & 1778.7  & 1741.6  \\ 
   &  & MIRROR & 0.413 & 1456.1 & 1367.5  \\ 
   &  & \textbf{RLB-MI (Ours)} & 0.659 & \textbf{1310.7} & \textbf{1214.7}  \\ 
   \cmidrule{2-6}
   &  Label-only  & BREP-MI & 0.581 & 1347.4 & 1256.5  \\ 
   \midrule
\multirow{6}{*}{\shortstack{ResNet-152 \\ (91\%)}} & \multirow{2}{*}{White-box} & GMI & 0.300 &  1594.1 & 1503.5 \\ 
   &  & KED-MI & 0.765 & 1277.3 & 1184.6 \\ 
   \cmidrule{2-6}
   &  \multirow{3}{*}{Black-box}& LB-MI & 0.041 & 1800.9  & 1735.7  \\ 
   &  & MIRROR & 0.570 & 1360.7 & 1263.8  \\ 
   &  & \textbf{RLB-MI (Ours)} & \textbf{0.804} & \textbf{1217.9} & \textbf{1108.2}  \\ 
   \cmidrule{2-6}
   &  Label-only & BREP-MI & 0.729  & 1277.5 & 1180.4 \\ 
   \midrule
\multirow{6}{*}{\shortstack{Face.evoLVe \\ (89\%)}} & \multirow{2}{*}{White-box} & GMI & 0.254  &  1628.6 & 1541.7 \\ 
   &  & KED-MI & 0.741  & 1350.8 & 1261.6 \\ 
   \cmidrule{2-6}
   &  \multirow{3}{*}{Black-box}& LB-MI & 0.111  & 1776.4  & 1729.1  \\ 
   &  & MIRROR & 0.530 & 1379.7 & 1280.1  \\ 
   &  & \textbf{RLB-MI (Ours)} & \textbf{0.793}  & \textbf{1225.6} & \textbf{1112.1} \\ 
   \cmidrule{2-6}
   &  Label-only  & BREP-MI & 0.721  & 1267.3 & 1164.0  \\ 
   \bottomrule
\end{tabular}
}
\end{scriptsize}
\caption{Attack performance of the model inversion attacks on target models with various structures trained on CelebA with its test accuracy in parenthesis.} 
\label{result-table-models}
\end{table} 

\textbf{Performance evaluation on various models.} 
\Cref{result-table-models} shows the evaluation results of RLB-MI and the baselines on three models, VGG16, ResNet-152, and Face.evoLVe, trained with CelebA. The test accuracy of each target model is 88\%, 91\%, and 89\%, respectively. 
Although the existing black-box model inversion attacks, LB-MI and MIRROR, have access to soft labels, they showed significantly lower performance than BREP-MI, which is a label-only model inversion attack method. However, the proposed black-box attack, RLB-MI, greatly outperformed BREP-MI by appropriately utilizing the information from confidence scores.
In addition, our attack surpassed the state-of-the-art white-box model inversion attack, KED-MI, in the case of ResNet-152 and Face.evoLVe, despite the inability to access gradient information. Even in the case of VGG16, it is shown that the images reconstructed by RLB-MI capture the informative features of the target classes by having the lowest K-nearest neighbor distance and feature distance from real samples.
It also can be seen that higher prediction performance of the target classifier leads to increased attack performance.
This result makes sense as the classifier
with better performance contains more accurate and critical
information about the features of training data.
We present qualitative evaluation results in \Cref{result-image} by providing the real samples and attack images generated by the baselines and our method, as the evaluation classifier may not completely represent human judgment.

\textbf{Performance evaluation on various datasets.} We measure the performance of the attacks on Face.evoLVe models trained with PubFig83, FaceScrub, and CelebA. The test accuracy of each target model is 96\%, 97\%, and 89\%, respectively. 
\Cref{result-table-datasets} shows that the attack accuracy of RLB-MI outperforms all baselines. Although LB-MI showed feature distance that is competitive with other methods, it obtained very low attack accuracy. This is caused by the limitation of the ability of the LB-MI's inversion model to learn the general prior from the public dataset.

\begin{table}[t]
\centering
\begin{scriptsize}
\resizebox{\columnwidth}{!}{
\begin{tabular}{c|c|c|rrr}
  \toprule
 Dataset & Type & Method & Attack Acc  & KNN Dist & Feat Dist \\ 
  \midrule[0.7pt]
\multirow{6}{*}{\shortstack{PubFig83 \\ (96\%)}} & \multirow{2}{*}{White-box} & GMI & 0.100 &  3037.2 & 3307.6 \\ 
   &  & KED-MI & 0.300 & 2205.0 & 2418.8 \\ 
   \cmidrule{2-6}
   &  \multirow{3}{*}{Black-box}& LB-MI & 0.000 & 2109.8  & 2380.4 \\ 
   &  & MIRROR & 0.280 & 2661.6 & 2869.4  \\ 
   &  & \textbf{RLB-MI (Ours)} & \textbf{0.420} & 2113.0 & 2257.2   \\ 
   \cmidrule{2-6}
   &  Label-only  & BREP-MI & 0.400  & \textbf{2073.8} & \textbf{2254.3}  \\ 
   \midrule
\multirow{6}{*}{\shortstack{FaceScrub \\ (97\%)}} & \multirow{2}{*}{White-box} & GMI & 0.180  &  2364.2 & 2455.4 \\ 
   &  & KED-MI & 0.400 & 2015.3 & 2081.9 \\ 
   \cmidrule{2-6}
   &  \multirow{3}{*}{Black-box}& LB-MI & 0.010  & 2266.3 & 2359.8  \\ 
   &  & MIRROR & 0.325 & 1995.1 & 2053.9  \\ 
   &  & \textbf{RLB-MI (Ours)} & \textbf{0.475}  & \textbf{1893.6} & \textbf{1945.3} \\ 
   \cmidrule{2-6}
   &  Label-only & BREP-MI & 0.325  & 2067.5 & 2136.3 \\ 
   \midrule
\multirow{6}{*}{\shortstack{CelebA \\ (89\%)}} & \multirow{2}{*}{White-box} & GMI & 0.254  &  1628.6 & 1541.7 \\ 
   &  & KED-MI & 0.741  & 1350.8 & 1261.6 \\ 
   \cmidrule{2-6}
   &  \multirow{3}{*}{Black-box}& LB-MI & 0.111  & 1776.4  & 1729.1 \\ 
   &  & MIRROR & 0.530 & 1379.7 & 1280.1  \\ 
   &  & \textbf{RLB-MI (Ours)}  & \textbf{0.793}  & \textbf{1225.6} & \textbf{1112.1}  \\ 
   \cmidrule{2-6}
   &  Label-only  & BREP-MI & 0.721  & 1267.3 & 1164.0  \\ 
   \bottomrule
\end{tabular}
}
\end{scriptsize}
\caption{Attack performance of the model inversion attacks on target models trained with various datasets with its
test accuracy in parenthesis.}
\label{result-table-datasets}
\end{table}

\textbf{Effect of differences in public dataset.}
In most realistic scenarios, the chance of the public and private datasets being in the same distribution is scarce, thus we conduct experiments by training the generative model on the public dataset that is in a different distribution from the private dataset. 
We evaluate the attacks using generative models trained with Flickr-Faces-HQ Dataset (FFHQ) \cite{Karras_2019_CVPR}.
The target classifiers used in these experiments are Face.evoLVe trained with PubFig83, FaceScrub, and CelebA.
In the experiments using FFHQ as public datasets, model inversion attacks showed decreased attack performance. However, even in the presence of distribution shifts between public and private datasets, our attack still achieves state-of-the-art attack performance as shown in \Cref{result-table-cross}.
We think that the performance decrease may have occurred because the way faces were aligned or cropped was different for each dataset and the gender or age distribution included in each dataset was different. When taking a picture, the differences in the image distribution created by the light condition or the background may also affect the attack performance.

\subsection{Experiments on the Attack Configurations}

\begin{figure}[t]
\centering
\begin{subfigure}[b]{0.23\textwidth}
\centering
\includegraphics[width=\textwidth]{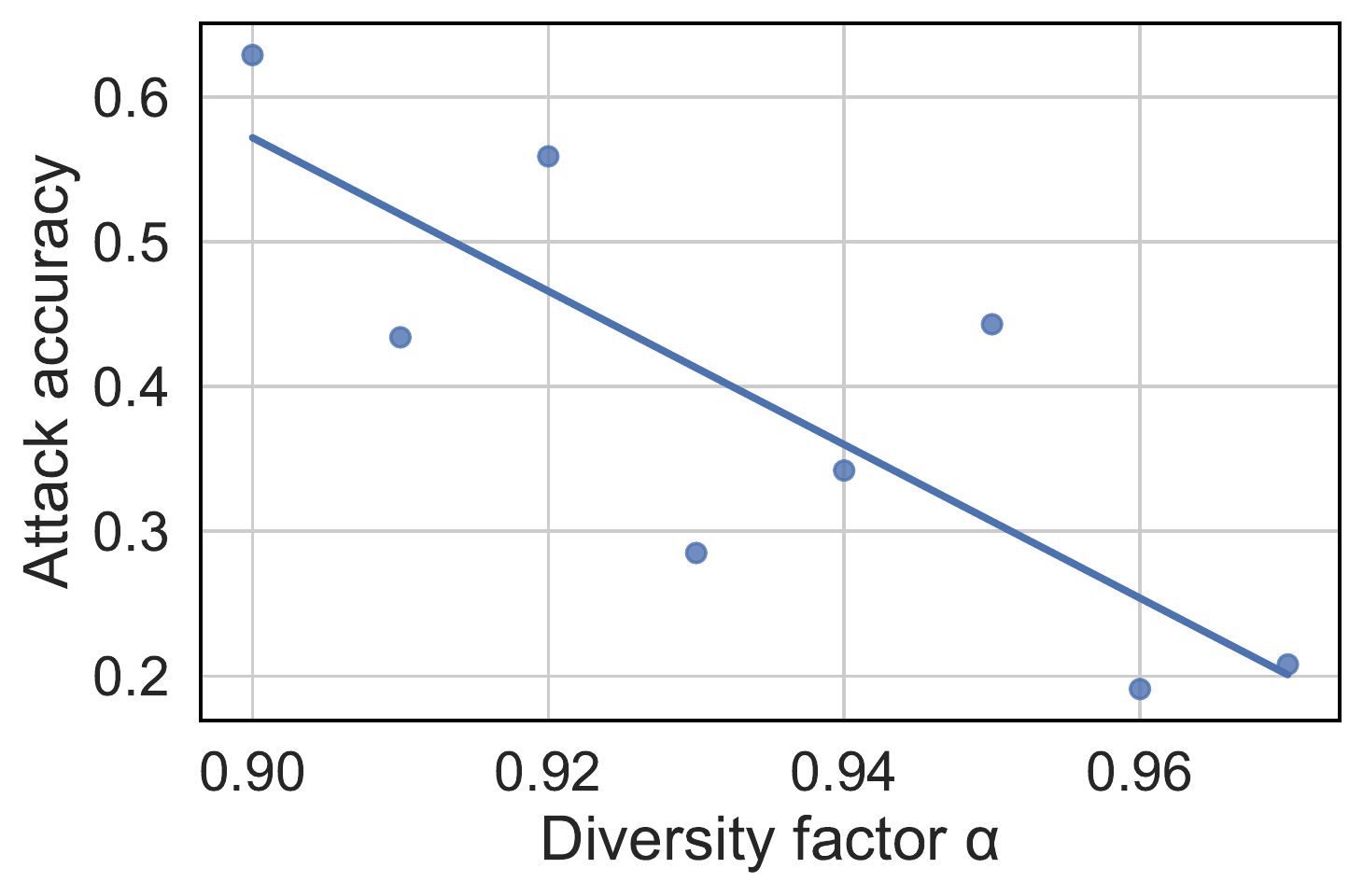}
\caption{Attack accuracy}
\end{subfigure}
\hfill
\begin{subfigure}[b]{0.23\textwidth}
\centering
\includegraphics[width=\textwidth]{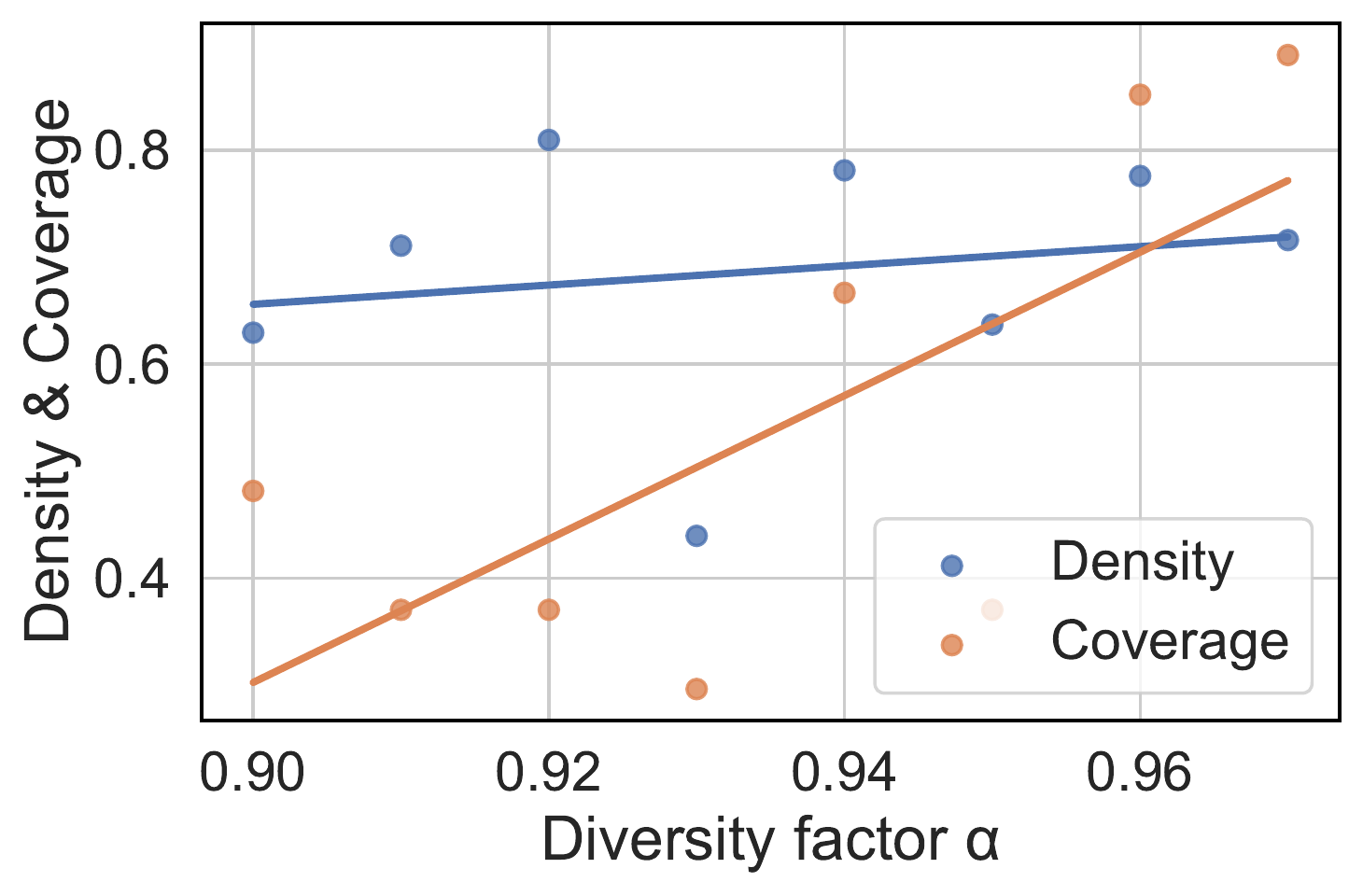}
\caption{Density \& Coverage}
\label{DandC}
\end{subfigure}
\caption{These graphs show the trade-off of our attack between accuracy and diversity. The graphs plot (a) the attack accuracy, (b) the density and coverage for the $\alpha$ values.}
\label{diversity-graph}
\end{figure}

\begin{figure}[t]
\centering
\begin{subfigure}[b]{0.236\textwidth}
\centering
\includegraphics[width=\textwidth]{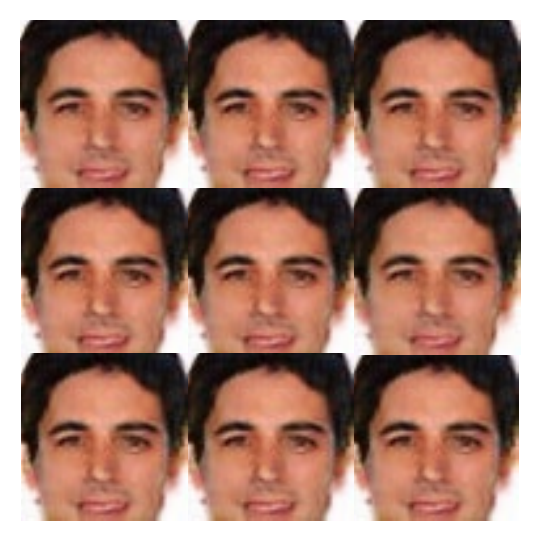}
\caption{$\alpha=0.00$}
\end{subfigure}
\hfill
\begin{subfigure}[b]{0.236\textwidth}
\centering
\includegraphics[width=\textwidth]{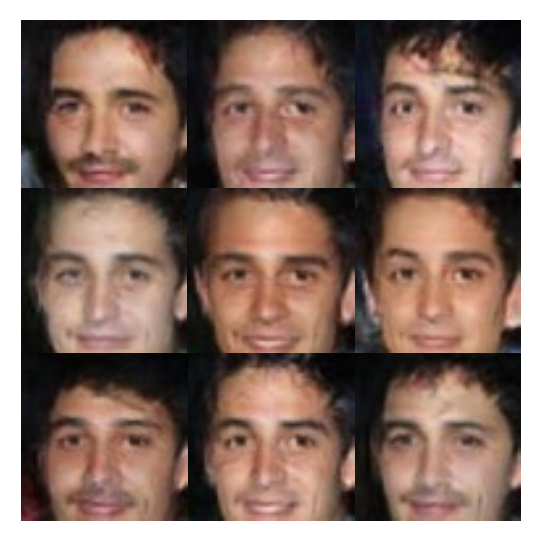}
\caption{$\alpha=0.97$}
\end{subfigure}
\caption{The images are reconstructed by RLB-MI with (a) $\alpha=0.00$ and (b) $\alpha=0.97$, respectively.}
\label{diversity-image}
\end{figure}

\textbf{Trade-off between accuracy and diversity.}
We measure the attack accuracy and diversity of trained agents depending on the changes in diversity factor $\alpha$. We train agents for various values of $\alpha$ and generate 1,000 images from each agent for a specific identity. To separately evaluate the fidelity and diversity of the generated images, we use Density and Coverage (D\&C) \cite{Baek_2021_ICCV} as metrics. The target classifier used in these experiments is Face.evoLVe trained with CelebA. \Cref{diversity-graph} shows the trade-off between attack accuracy and diversity. As alpha increases, the attack accuracy decreases, and the coverage that represents diversity increases. Also, it can be seen from \Cref{DandC} that the density is robust to changes in alpha, which means that agents generate images of steady fidelity regardless of changes in alpha. The images generated when $\alpha$ is 0.00 and 0.97 are visualized in \Cref{diversity-image}. 

\begin{table}[t]
\centering
\begin{scriptsize}
\resizebox{\columnwidth}{!}{
\begin{tabular}{c|c|c|rrr}
  \toprule
 Public$\rightarrow$Private & Type & Method & Attack Acc & KNN Dist & Feat Dist \\ 
  \midrule[0.7pt]
\multirow{6}{*}{FFHQ$\rightarrow$PubFig83} & \multirow{2}{*}{White-box} & GMI & 0.080 & 2913.0 & 3126.9 \\ 
   &  & KED-MI & 0.300  & 2587.5 & 2794.4  \\ 
   \cmidrule{2-6}
   &  \multirow{3}{*}{Black-box}& LB-MI & 0.040  & 2554.8  & 2804.6  \\ 
   &  & MIRROR & 0.260 & 2419.9 & 2621.8  \\ 
   &  & \textbf{RLB-MI (Ours)} & \textbf{0.400}  & \textbf{2171.9} & \textbf{2352.9}  \\ 
   \cmidrule{2-6}
   &  Label-only  & BREP-MI & 0.360 & 2331.0 & 2491.2  \\ 
   \midrule
\multirow{6}{*}{FFHQ$\rightarrow$FaceScrub} & \multirow{2}{*}{White-box} & GMI & 0.160  & 2583.2 & 2683.5 \\ 
   &  & KED-MI & 0.285  & 2382.3 & 2483.6 \\ 
   \cmidrule{2-6}
   &  \multirow{3}{*}{Black-box}& LB-MI & 0.005 & 2807.7  & 2952.9  \\ 
   &  & MIRROR & 0.250 & 2275.9 & 2366.0  \\ 
   &  & \textbf{RLB-MI (Ours)} & \textbf{0.385}  & \textbf{2204.1} & \textbf{2278.5}  \\ 
   \cmidrule{2-6}
   &  Label-only & BREP-MI & 0.260 & 2259.9 & 2342.4 \\ 
   \midrule
\multirow{6}{*}{FFHQ$\rightarrow$CelebA} & \multirow{2}{*}{White-box} & GMI & 0.100  & 1789.8 & 1692.5 \\ 
   &  & KED-MI & 0.427  & 1518.4 & 1421.6 \\ 
   \cmidrule{2-6}
   &  \multirow{3}{*}{Black-box}& LB-MI & 0.047  & 1930.5 & 1855.2  \\ 
   &  & MIRROR & 0.266 & 1608.4 & 1511.6  \\ 
   &  & \textbf{RLB-MI (Ours)} & \textbf{0.433} & \textbf{1481.9} & \textbf{1361.8} \\ 
   \cmidrule{2-6}
   &  Label-only  & BREP-MI & 0.410  & 1498.6 & 1386.2  \\ 
   \bottomrule
\end{tabular}
}
\end{scriptsize}
\caption{Attack performance of the model inversion attacks when the distribution of public data differs from the distribution of private
data.} 
\label{result-table-cross}
\end{table}

\begin{figure}[t]
\begin{center}
\centering
\begin{subfigure}[b]{0.23\textwidth}
\centering
\includegraphics[width=\textwidth]{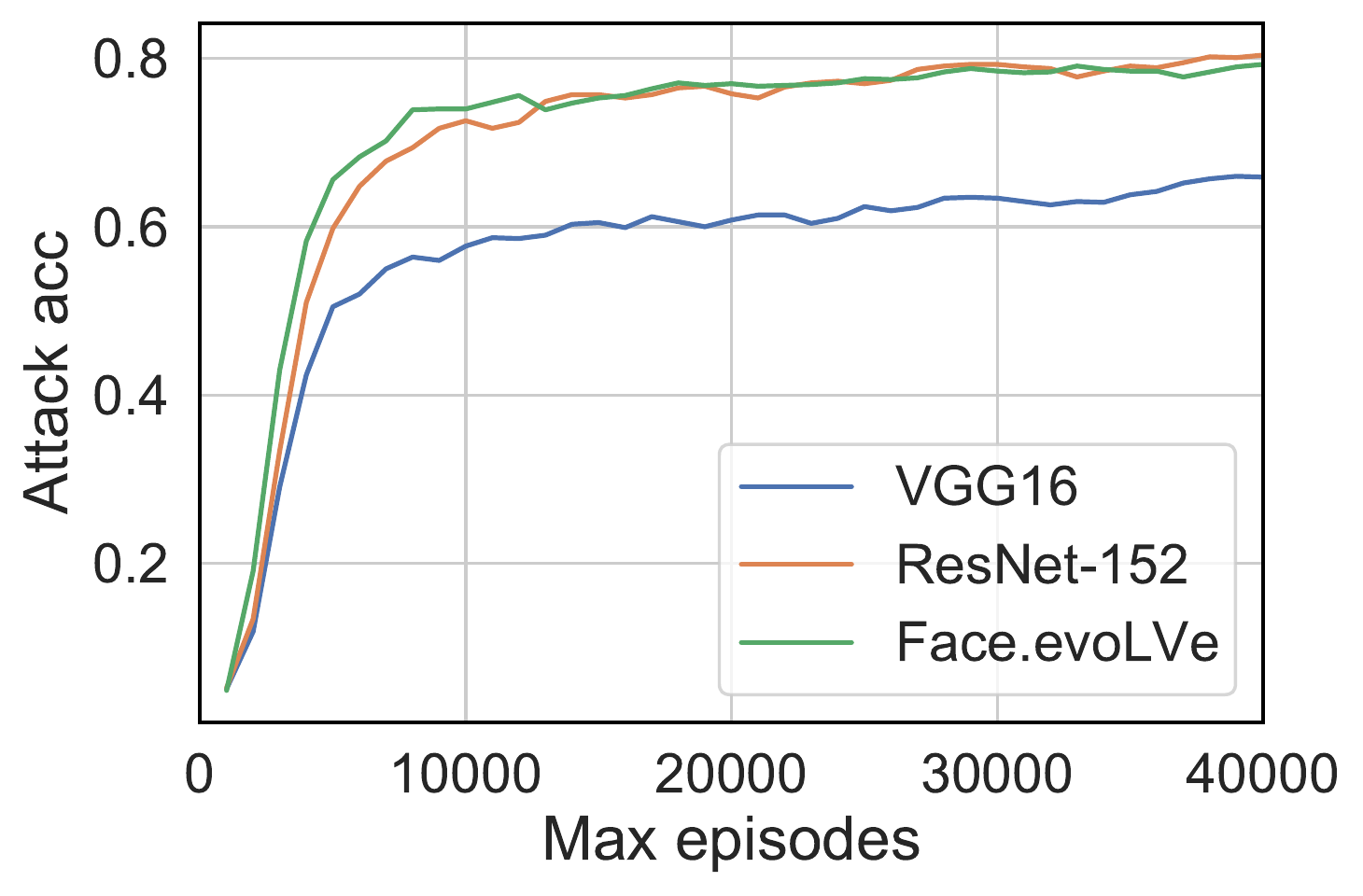}
\caption{Models}
\label{episode-models}
\end{subfigure}
\hfill
\begin{subfigure}[b]{0.23\textwidth}
\centering
\includegraphics[width=\textwidth]{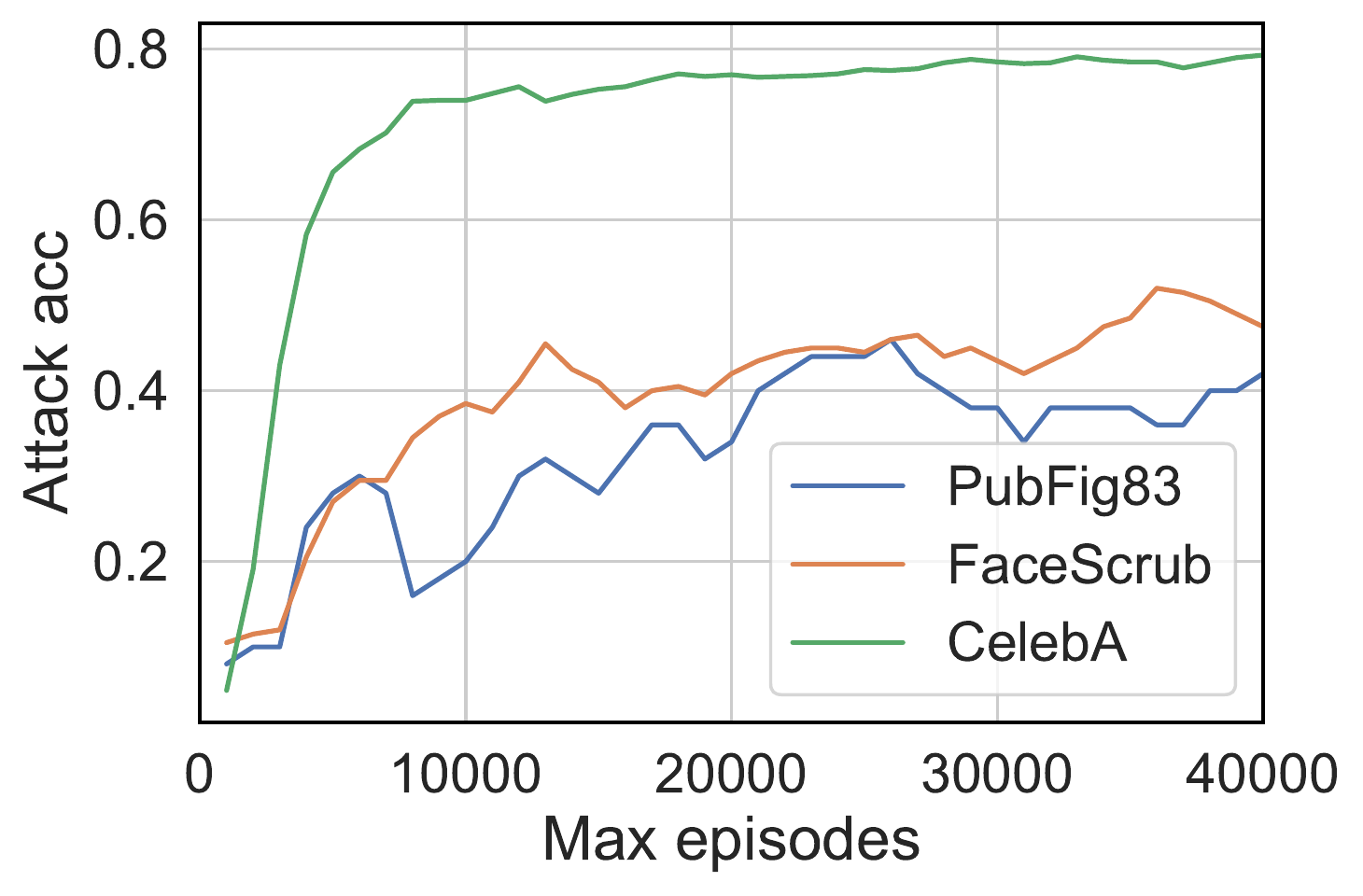}
\caption{Datasets}
\label{episode-datasets}
\end{subfigure}
\caption{Attack accuracy of RLB-MI depending on the maximum number of episodes. For (a), the dataset is set as CelebA and the experimental results for several target models are reported, and for (b), the target model is set as Face.evoLVe, and the experimental results for several datasets are reported.}
\label{episode-graph}
\end{center}
\end{figure}

\textbf{Performance tendency depending on the maximum number of episodes.} To observe the change in attack performance depending on the maximum number of episodes, we report the attack accuracy by increasing the maximum episode in the step of 1,000 from 1,000 to 40,000. The experimental results can be checked in \Cref{episode-graph}. In \Cref{episode-models}, we plot results for various target models trained with CelebA, and in \Cref{episode-datasets} we plot results for Face.evoLVe models trained with various datasets.
The attack performance per maximum episode increased rapidly at the beginning and then saturated. Even when the structure of the target classifier is changed, there was no significant difference in the tendency of the attack accuracy when the dataset is the same. In addition, the smaller the number of the target classes of the dataset, the earlier the saturation point appeared, and after that, the attack accuracy decreased due to overfitting. For PubFig83 with 50 target classes, the saturation point appeared at 25,000 episodes, and for FaceScrub with 200 target classes, the saturation point appeared at 36,000 episodes. Therefore, it is important to set proper maximum number of episodes depending on the target datasets.

\begin{figure}[ht]
\begin{center}
\centerline{
\includegraphics[width=1.0\columnwidth]{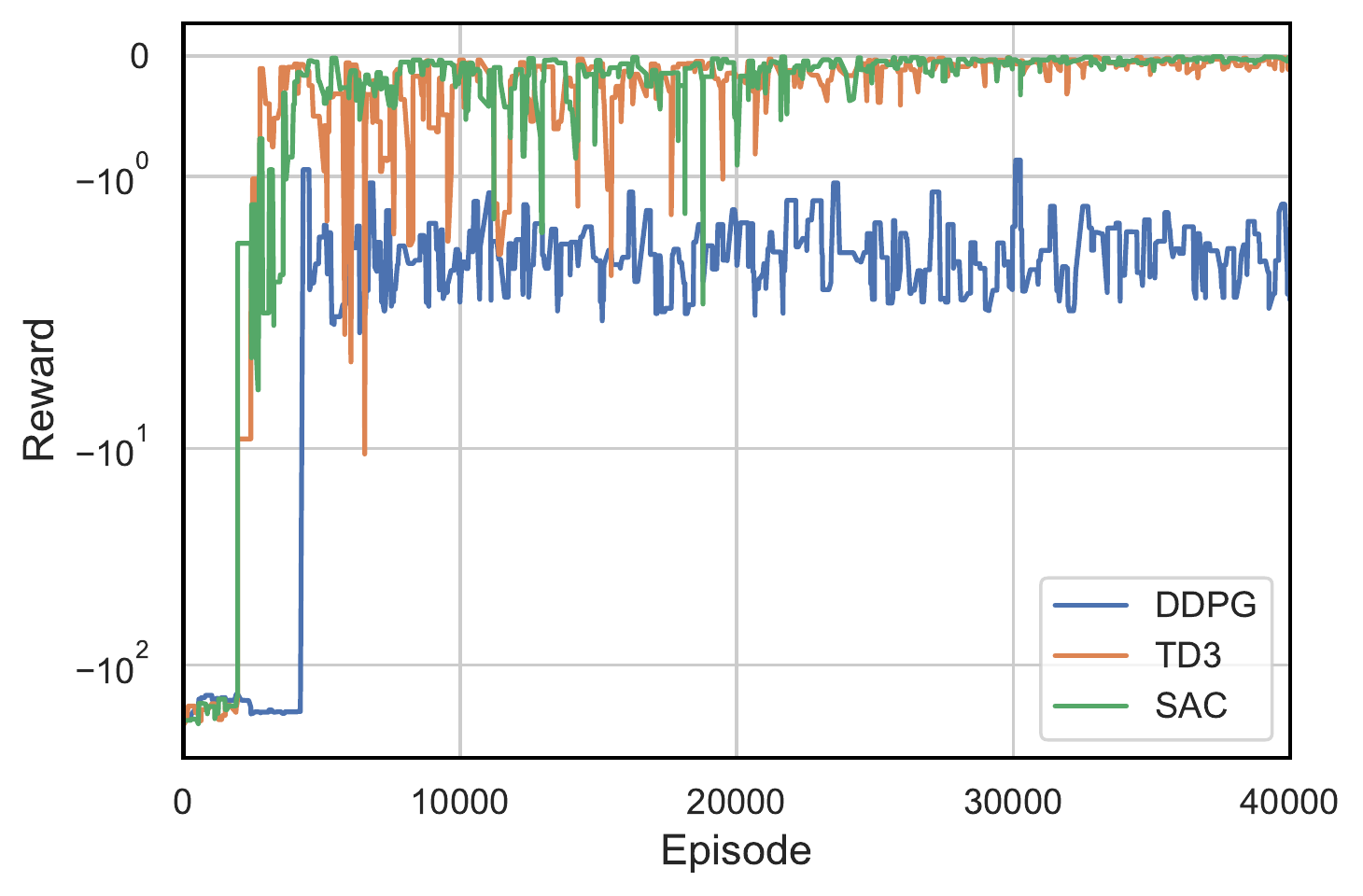}
}
\caption{The total reward for each episode when using various RL agents. The plotted values are passed through a one-dimensional maximum filter of size 5 for visibility.}
\label{agents-graph}
\end{center}
\end{figure}

\textbf{Experiments with various RL agents.} To determine the effect of the RL agents, we experiment with DDPG \cite{DBLP:journals/corr/LillicrapHPHETS15} and TD3 \cite{pmlr-v80-fujimoto18a} instead of SAC for the latent space search. The target classifier used in this experiment is Face.evoLVe trained with CelebA.  The results in \Cref{result-table-agent} show that when SAC is used in our attack, it achieves a performance significantly higher than that of DDPG and TD3. We consider the cause of these results as SAC's robustness to noisy and complex environments. In our attack, a high-dimensional random latent vector is newly given every episode, and the confidence score of the target identity changes sensitively to such a latent vector. Therefore, a reinforcement learning algorithm that is robust against noisy and complex environments like SAC is desirable. 
In addition, we show the change in the reward of each RL agent in \Cref{agents-graph}.

\begin{table}[t]
\centering
\begin{footnotesize}
\begin{tabular}{c|rrrr}
  \toprule
  Agent & Attack Acc  & KNN Dist & Feat Dist \\ 
  \midrule[0.7pt]
   DDPG & 0.380 & 1532.2 & 1445.9 \\ 
   TD3 &  0.677 & 1325.5 & 1215.0  \\ 
   SAC & \textbf{0.793}  & \textbf{1225.6} & \textbf{1112.1}  \\ 
   \bottomrule
\end{tabular}
\end{footnotesize}
\caption{Attack performance with various RL agents.} 
\label{result-table-agent}
\end{table} 

\section{Ethical Considerations}

There may be negative social impacts such as infringement of personal information that needs protection if the proposed black-box model inversion attack is abused by malicious users.
However, revealing the vulnerabilities of the current system is indispensable for the development of security.
Through this study, we raise awareness about the privacy issues of machine learning and urge the community to develop algorithms or systems to defend against suggested vulnerabilities. 
We believe that our work will have a positive impact that outweighs the risks in terms of security.

\section{Conclusion}

We propose a new black-box model inversion attack using a GAN based on reinforcement learning. We formulate the latent space exploration as an MDP problem, and train a reinforcement learning agent to solve the MDP, even in the absence of information about the target model such as weights and gradients.
The proposed attack addressed the problems with the previous black-box attacks.
In addition, the experimental results show that our attack successfully reconstructs the private data of the target model. Our attack outperforms not only state-of-the-art black-box attacks but also all the other methods including the white-box and label-only attacks.
We hope that this study will invigorate studies on the black-box model inversion attacks and defenses.

\section*{Acknowledgments}
This work was supported by Institute of Information \& communications Technology Planning \& Evaluation (IITP) grant funded by the Korea government(MSIT). (NO.2022-0-00184, Development and Study of AI Technologies to Inexpensively Conform to Evolving Policy on Ethics)
{\small
\bibliographystyle{ieee_fullname}
\bibliography{CameraReady.bbl}
}

\end{document}